\documentclass[a4paper, 10pt, twocolumn, twoside]{article}

\usepackage{ISARC}

\usepackage{lscape}
\usepackage{hologo}
\usepackage{multirow}
\usepackage{makecell}

\begin{document}

\linespread{0.5}

\title{Training and Simulation of Quadrupedal Robot in Adaptive Stair Climbing for Indoor Firefighting: An End-to-End Reinforcement Learning Approach}

\author{Baixiao Huang$^{1 \star}$  Baiyu Huang$^{1 \star}$  and Yu Hou$^{2 \dag}$}

\affiliation{
$^1$Independent Researcher\\
$^2$Department of Construction Management, Western New England University\\
$^{\star}$These authors contribute equally to this work\\
$^{\dag}$Corresponding author
}

\email{
\href{mailto:e.author1@aa.bb.edu}{baixiaohuang1@gmail.com}, 
\href{mailto:e.author1@aa.bb.edu}{baiyuhuang2@gmail.com},
\href{mailto:e.author1@aa.bb.edu}{yu.hou@wne.edu}
}

\maketitle 
\thispagestyle{fancy} 
\pagestyle{fancy}
\fancyhf{}              
\fancyfoot[C]{\thepage}

\begin{abstract}
Quadruped robots are used for primary searches during the early stages of indoor fires. A typical primary search involves quickly and thoroughly looking for victims under hazardous conditions and monitoring flammable materials. However, situational awareness in complex indoor environments and rapid stair climbing across different staircases remain the main challenges for robot-assisted primary searches. In this project, we designed a two-stage end-to-end deep reinforcement learning (RL) approach to optimize both navigation and locomotion. In the first stage, the quadrupeds, \textit{Unitree Go2}, were trained to climb stairs in \textit{Isaac Lab}’s pyramid-stair terrain. In the second stage, the quadrupeds were trained to climb various realistic indoor staircases in the \textit{Isaac Lab} engine, with the learned policy transferred from the previous stage. These indoor staircases are straight, L-shaped, and spiral, to support climbing tasks in complex environments. This project explores how to balance navigation and locomotion and how end-to-end RL methods can enable quadrupeds to adapt to different stair shapes. Our main contributions are: (1) A two-stage end-to-end RL framework that transfers stair-climbing skills from abstract pyramid terrain to realistic indoor stair topologies. (2) A centerline-based navigation formulation that enables unified learning of navigation and locomotion without hierarchical planning. (3) Demonstration of policy generalization across diverse staircases using only local height-map perception. (4) An empirical analysis of success, efficiency, and failure modes under increasing stair difficulty.
\end{abstract}


\section{Introduction}
\label{sec:Introduction}

During indoor fires, the primary search phase involves a rapid, meticulous survey of the environment to locate potential victims, often under hazardous conditions. This critical task also involves continuous monitoring for highly flammable and combustible materials that could pose an immediate threat to first responders. With the development of artificial intelligence (AI) and advanced robotics, quadrupedal robots, commonly referred to as robot dogs, have become a vital tool for primary search tasks. These quadrupeds possess several distinct advantages that make them superior to other robotic platforms in this specialized application. Their cost-effectiveness is a significant factor, making them more accessible for deployment by various emergency services. Moreover, their superior adaptability allows them to traverse uneven terrain and navigate cluttered indoor spaces that would immobilize wheeled or tracked robots. Crucially, their inherent stability, derived from their four-legged gait, provides a robust platform for sensor deployment and reliable movement across debris-strewn floors. Finally, researchers also use indoor drones for primary search, but they cannot handle obstacles as effectively as quadrupeds. For example, mounting manipulator arms on robot dogs allows them to open doors or clear barriers during primary search.

Despite these strengths, the use of robot dogs in primary searches is still constrained by two major technical hurdles. The first is achieving comprehensive situational awareness within the complex and dynamic indoor fire environment. This challenge involves overcoming visual obstructions caused by smoke, accurately mapping three-dimensional space, and interpreting sensor data (such as thermal readings) to build a clear, actionable picture of the environment for human commanders. Researchers are studying the application of building information modeling (BIM) to support robot dogs’ situational awareness. The second challenge is developing robust, high-speed stair-climbing capabilities. Given the architectural variety of buildings, the robots must be able to rapidly and reliably ascend and descend a multitude of staircase types. These staircase types include regular, spiral, and damaged steps to facilitate a rapid and comprehensive primary search across all floors. Overcoming this challenge is essential for maximizing the effectiveness of robot-assisted primary searches. 

In this project, we explored an innovative approach to solving the quadruped's stair-climbing challenges. This approach presents a two-stage end-to-end deep reinforcement learning (RL) framework to optimize both navigation and locomotion. We trained and tested our algorithm with the NVIDIA \textit{Isaac Lab} platform and implemented our approach on a widely used quadruped robot prototype, \textit{Unitree Go2}. In the first stage, the quadruped was trained to climb stairs in \textit{Isaac Lab}’s pyramid-stair terrain. In the second stage, dogs were trained to climb various realistic indoor staircases in the \textit{Isaac Lab} engine, using the policy learned in the previous stage. There are many types of indoor staircases in residential and commercial houses. In order to test our approach, we conducted experiments on three typical indoor staircases, including straight, L-shaped, and spiral stairs, to support climbing tasks in complex environments. 

This project explores how to balance navigation and locomotion and how end-to-end RL methods can enable the robot dogs to adapt to different stair shapes while maintaining high agility.

\section{Related work}
\label{sec:RelatedWork}

\subsection{Robots for Indoor Primary Search}
In Firefighting, primary search requires an initial and rapid search to locate and remove victims quickly and safely while the fire is still active and conditions remain dangerous. Researchers have implemented robots for primary search. Gelfert et al. \cite{Gelfert2022} and Talavera et al. \cite{Talavera2023} developed a mobile robot concept to assist firefighters in locating victims in smoky indoor apartment fires using an Unmanned Ground Vehicle (UGV), but their UGVs are tracked-based and consider only one floor in a fire scenario. Pritzl et al. \cite{pritzl2021} and Han et al. \cite{Han2025} used an Unmanned Aerial Vehicle (UAV) to autonomously enter a target building through an open window in a firefighting scenario. However, UAVs cannot handle obstacles and can only access fire zones through openings, such as open doors and windows. Based on this discussion, quadruped robots offer a broader range of indoor firefighting use cases than existing UGVs and UAVs.

\subsection{Quadruped Robot Prototypes and Training Platforms}

There are numerous representative platforms and reference quadruped robot prototypes, including those from Boston Dynamics \cite{Website_Boston}, Unitree Robotics \cite{Website_Unitree}, Ghost Robotics \cite{Website_Ghost}, and ANYbotics/ANYmal. They have been used for building-environment inspection, disaster response, search and rescue, and military reconnaissance \cite{Website_Ghost}. NVIDIA \textit{Isaac Lab} is a widely used open-source robotics learning platform for developing RL algorithms and deploying Sim-to-Real workflows.

\subsection{Quadruped Robot Climbing Tasks}
Qi et al. \cite{Qi2021} and Liang et al. \cite{Liang2022} used a model-based framework to explore the stair-climbing algorithms with perception, planning, and control structures. However, the stair environments are simplified to straight shapes. Vogel et al. \cite{Vogel2025} explored ladder climbing (equivalent to a 90° stair) using a custom hook-shaped end effector, but their method cannot be generalized to other staircases due to its reliance on specialized hardware. Hoeller et al. \cite{Hoeller2024} utilized ANYmal to perform agile navigation in a complex environment. They decomposed the problem into perception, locomotion, and navigation modules. In the locomotion module, the trained quadrupedal robot walks, jumps, and climbs. However, their testing environments are static. Kim et al. \cite{Kim2025} proposed a hierarchical navigation framework with a low-level tracker controller and a high-level foothold planner. The framework enables quadrupedal robots to traverse discrete terrain at high speed. The hierarchical framework could potentially increase system complexity and exclude globally optimal behaviors. 

In summary, future quadruped robot climbing tasks need to be investigated to develop a general end-to-end framework for various stair shapes to address the complex staircases. The end-to-end methods are simple to design, fast to execute, and naturally handle perception-control feedback. Additionally, training robots to be familiar with various stair shapes helps them conduct indoor fire primary search effectively.

\begin{figure*}[!htb]
    \centering
    \includegraphics[width=0.9\textwidth]{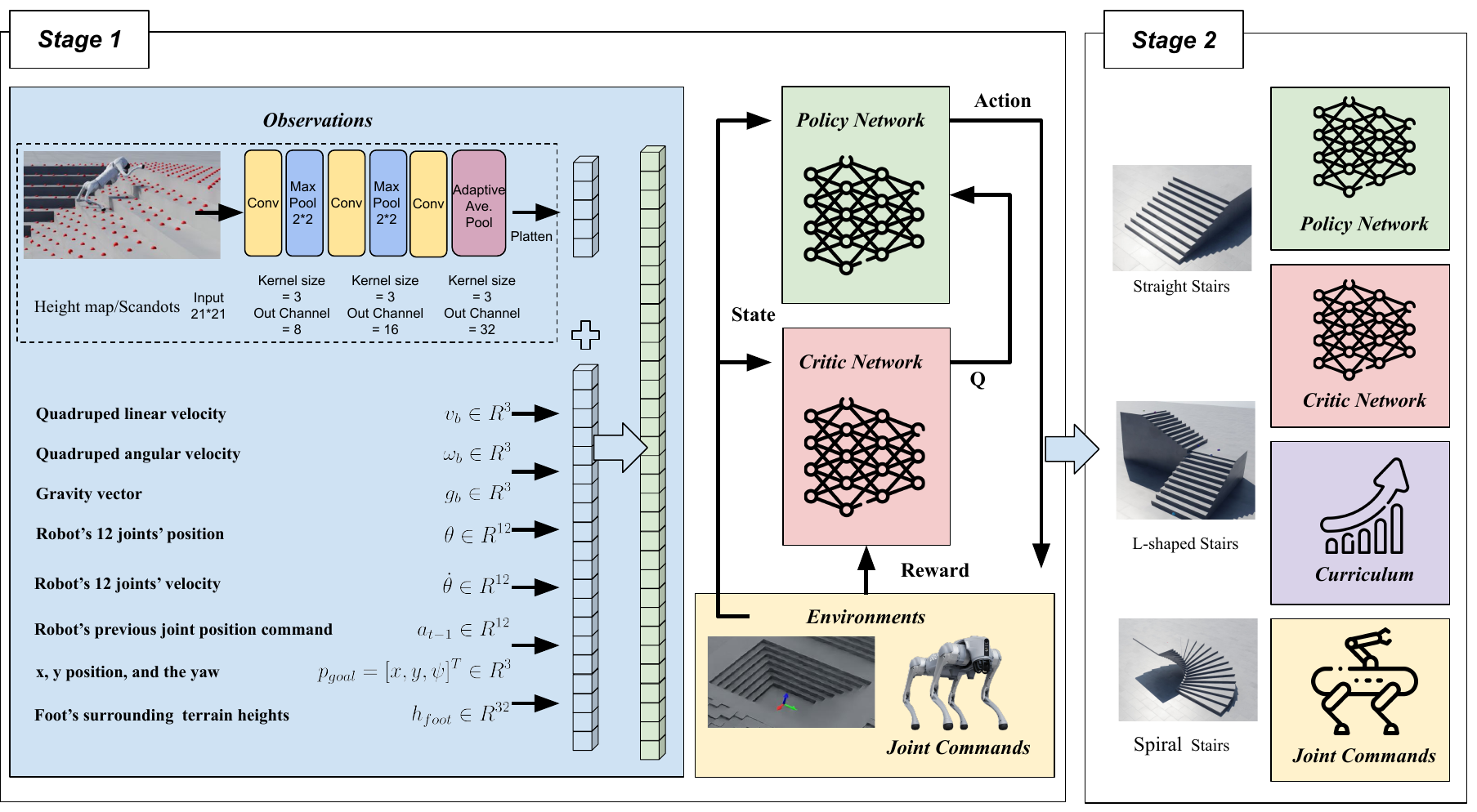}
    \caption{Two-Stage training}
    \label{fig_1}
\end{figure*}

\section{Methodology}
\label{sec:Methodology}

We trained a neural network policy to perform both the navigation and locomotion tasks on the stairs terrain. The policy takes proprioceptive states, environmental states represented as a height map, and user-commanded goals as inputs/observations. It then outputs joint position commands as actions, and robot dogs’ Proportional-Derivative (PD) controllers convert joint position commands into torques applied to joint motors.

The neural network is trained by using deep RL with the on-policy algorithm Proximal Policy Optimization (PPO) \cite{schulman2017proximalpolicyoptimizationalgorithms} in the \textit{Isaac Lab} environment. The training framework consists of two stages. The first stage trains the policy on stairs terrain to acquire basic stair-climbing locomotion skills. As shown in Figure~\ref{fig_1}, the terrain, preprogrammed by \textit{Isaac Lab}, is a pyramid-stair pattern that tapers to a flat platform at its center. The robot is placed at the center of the staircase and tasked with navigating to a specific goal position and orientation. The second stage trains the policy on our various customized stair terrains, such as straight, L-shaped, and spiral stairs, to develop terrain-specific navigation skills in addition to basic locomotion skills. In the second stage, the robot is tasked with tracking the stair centerline as it ascends to the upper floor.

\subsection{Observations}
The inputs, or observations, to the neural network are defined as $o = \left [ v_{b}, \omega _{b}, g_{b}, \theta, \dot{\theta},a_{t-1}, p_{goal}, e \right ]$ , where $v_{b} \in \mathbb{R}^{3}$ and $\omega_{b} \in \mathbb{R}^{3}$ are the quadruped’s linear and angular velocity in the body frame, and $g_{b} \in \mathbb{R}^{3}$ is the gravity vector in the body frame. In the observation equation, $\theta \in \mathbb{R}^{12}$ and $\dot{\theta} \in \mathbb{R}^{12}$  are the position and velocity of 12 joints, $a_{t-1} \in \mathbb{R}^{12} $ is the 12 joint position command in the previous timestep passed to the PD controller, and $p_{goal} = \left [ x,y,\psi \right ]^{T} \in \mathbb{R}^{3}$ is the goal pose consisting of $x$, $y$ position, and the yaw of the goal pose. Lastly, $e = \left [ h_{foot},h_{grid} \right ]^{T}$ is an environmental state containing the heights of the terrain around the quadruped’s 4 feet with a space of  $h_{foot} \in \mathbb{R}^{32}$  and the heights of the terrain centering around the quadruped in a grid-like pattern with a space of $h_{grid} \in \mathbb{R}^{441}$ . The grid-like pattern has a 21x21 shape with a 0.2m resolution. These observations provide the policy with robot kinematic and proprioceptive states, as well as the terrain around the robot, which are necessary for the policy to produce navigation and locomotion strategies. Finally, the policy outputs joint action for the controller $a_{t} = f_{policy}(o)$, where $f_{policy}$ is the deep neural network, and $a_{t} \in \mathbb{R}^{3}$ is the joint position setpoint.

\subsection{Network architecture}
The neural network $ f_{policy}$ consists of a shallow convolutional neural network (CNN) encoder and a 3-layer multilayer perceptron (MLP). The CNN takes a 21-by-21 height map/ scandots as input and outputs a 128-dimensional feature vector representing terrain information to the MLP layer, along with other observations as described previously. The MLP then outputs the action, or 12 joint position command. The CNN consists of convolutional and pooling layers, while the MLP consists of 3 hidden layers with sizes 128, 128, and 64. 

The critic network has the same architecture as the policy network and shares the CNN encoder with it. The policy CNN encoder's output is fed into the critic MLP network. By sharing the CNN encoder, we remove the burden of training a separate encoder for terrain features. However, the critic's MLP has its own set of weights, which estimate long-term returns given the current robot and environmental states.

\subsection{Rewards and Penalties}

In this paper, we formulated the problem as a navigation task. The navigation task involves the robot reaching a specific goal position and orientation. In contrast, we did not formulate the task as a velocity-tracking problem because the robot may need to move at different velocities depending on the terrain. Therefore, by formulating navigation rewards, the robot can perform combined navigation and locomotion tasks with high agility. We divided all rewards into two groups based on their functions: task/navigation rewards and regularization rewards. The total rewards are thereby $r_{total} = r_{task} + \sum_{i=1}^{n} \lambda _{i} r_{reg}^{(i)}$, where $\lambda _{i}$ is the weight, and $r_{reg}^{(i)}$ represents the regularization reward and penalties.

\subsection{Task rewards}

In stage 1 training, we used a navigation reward to indicate how close the quadruped was to the goal. It is formulated as in Equation (~\ref{eq_0}).

\begin{equation}\label{eq_0}
r_{nav} = 1 - tanh (\frac{d_{goal}}{\sigma _{nav}})
\end{equation}

where $r_{nav}$ is the navigation reward, the $d_{goal}$ is the Euclidean distance from the robot to the goal, and $\sigma _{nav}$ is the range in which the reward takes effect. The larger the $\sigma _{nav}$ is, the further the influence of the reward is. The purpose of the $tanh$ function is to reshape the reward in the range of 0 and 1, while keeping the gradient of the reward smooth. We used two navigation rewards, the coarse navigation reward, $r_{nav, coarse}$, and the fine navigation rewards, $r_{nav, fine}$, to handle both long-distance and close-to-goal tracking. The $\sigma _{nav}$ values are 5.0 and 1.0, respectively, based on our training experience. 

In stage 2 training, the robot was presented with a different task, navigating up the stairs. Therefore, the coarse navigation reward was replaced with two rewards, a centering reward and a path reward. A centerline is the path situated in the middle of the stairs and running along them. We required the quadruped to stay close to this centerline and traverse it to reach the goal. This naturally leads to the centering reward and path reward, represented as Equation (~\ref{eq_1}) and Equation (~\ref{eq_2}).

\begin{equation}\label{eq_1}
r_{center} = 1 - tanh (\frac{d_{center}}{\sigma _{center}})
\end{equation}
\begin{equation}\label{eq_2}
r_{path} = (1 - tanh (\frac{d_{path}}{\sigma _{path}})) \times  (1 - tanh (\frac{d_{center}}{\sigma _{center}}))
\end{equation}

In Equations (~\ref{eq_1}) and (~\ref{eq_2}), $r_{center}$ and $r_{path}$ are the centering and path rewards, $d_{center}$ is the distance from the quadruped to the center line; $d_{path}$ is the distance to the goal along the centerline; $\sigma _{center} $and $\sigma _{path}$ determine the distance of influence similar to stage 1 navigation reward. Note that the path reward is conditioned on distance to the centerline, so that the reward can only be gained by traversing near the centerline. These two rewards encourage the robot to travel in the center of the stairs while navigating to the goal.

To track heading, we constructed a heading tracking penalty simply as, $r_{tracking}= - f_{angle}(\psi , \psi_{goal})$, where $\psi$ is the current quadruped heading or yaw, and $\psi_{goal}$ is the goal heading. $f_{angle}$ is a function that returns the angular distance between two angles. The negative sign indicates that this term is a penalty.

\begin{figure*}
    \centering
    \begin{subfigure}[b]{0.33\textwidth}
        \centering
        \includegraphics[width=5cm, height=3.8cm]{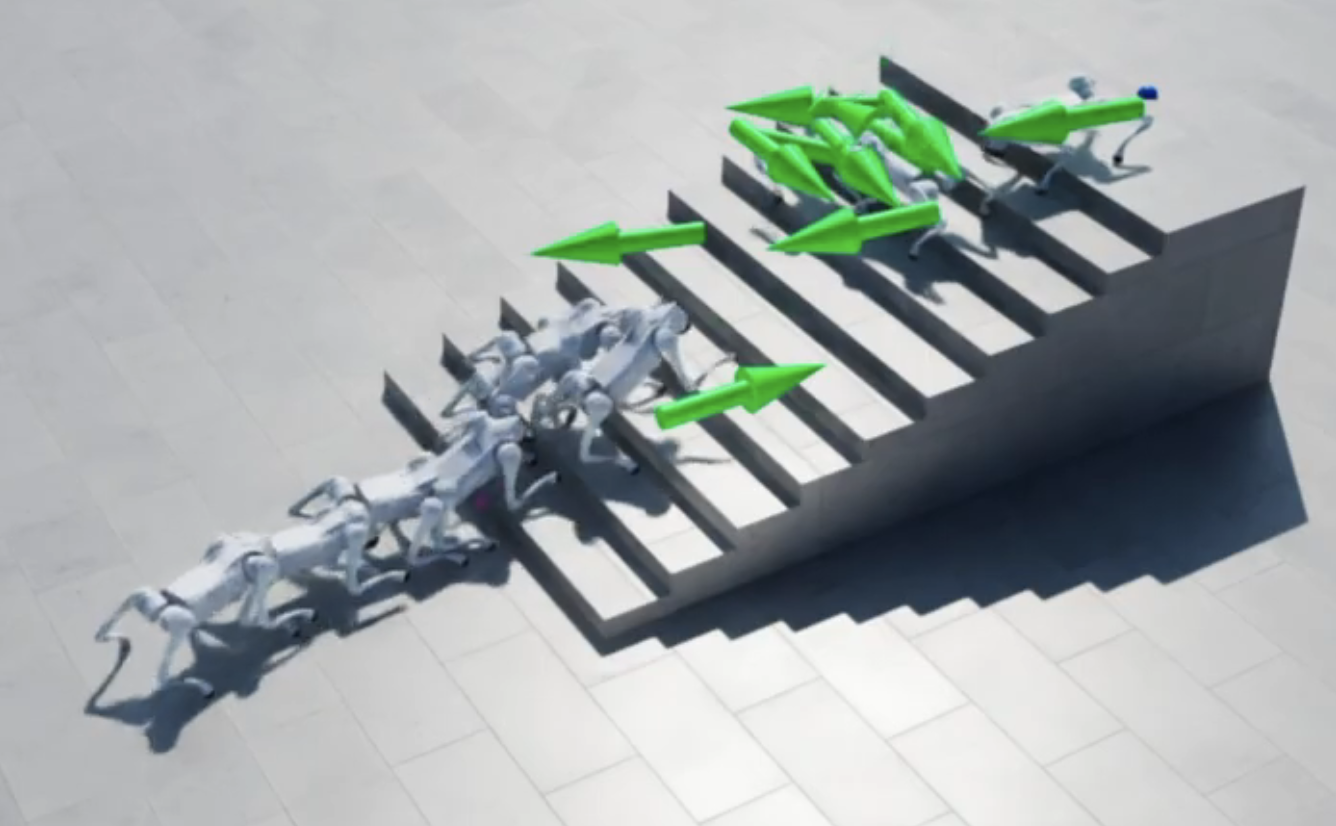}
        \caption{Robot Climbing Straight Stairs}
        \label{fig:sub1}
    \end{subfigure}
    \hfill 
    \begin{subfigure}[b]{0.33\textwidth}
        \centering
        \includegraphics[width=5cm, height=3.8cm]{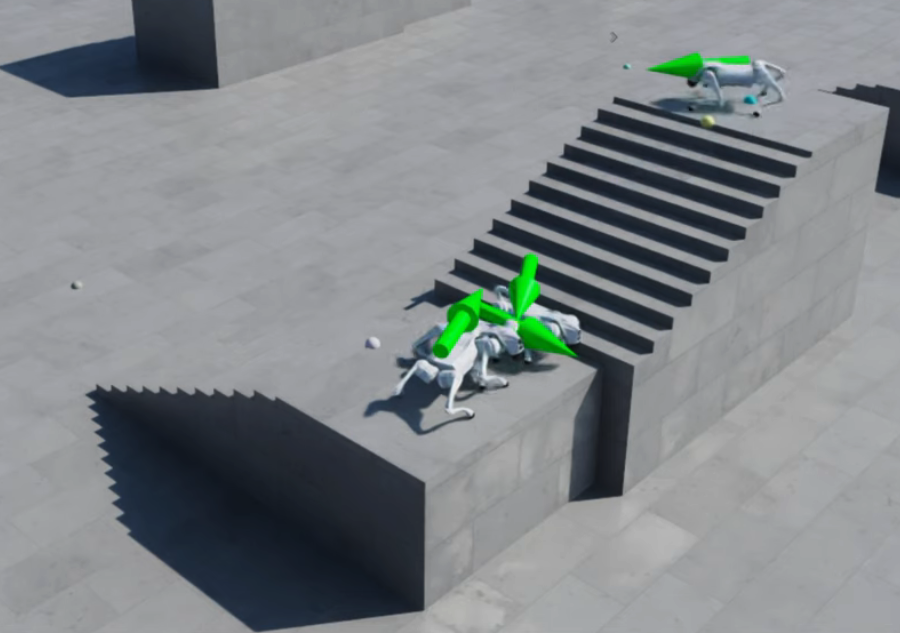}
        \caption{Robot Climbing L-Shaped Stairs}
        \label{fig:sub2}
    \end{subfigure}
        \begin{subfigure}[b]{0.33\textwidth}
        \centering
        \includegraphics[width=5cm, height=3.8cm]{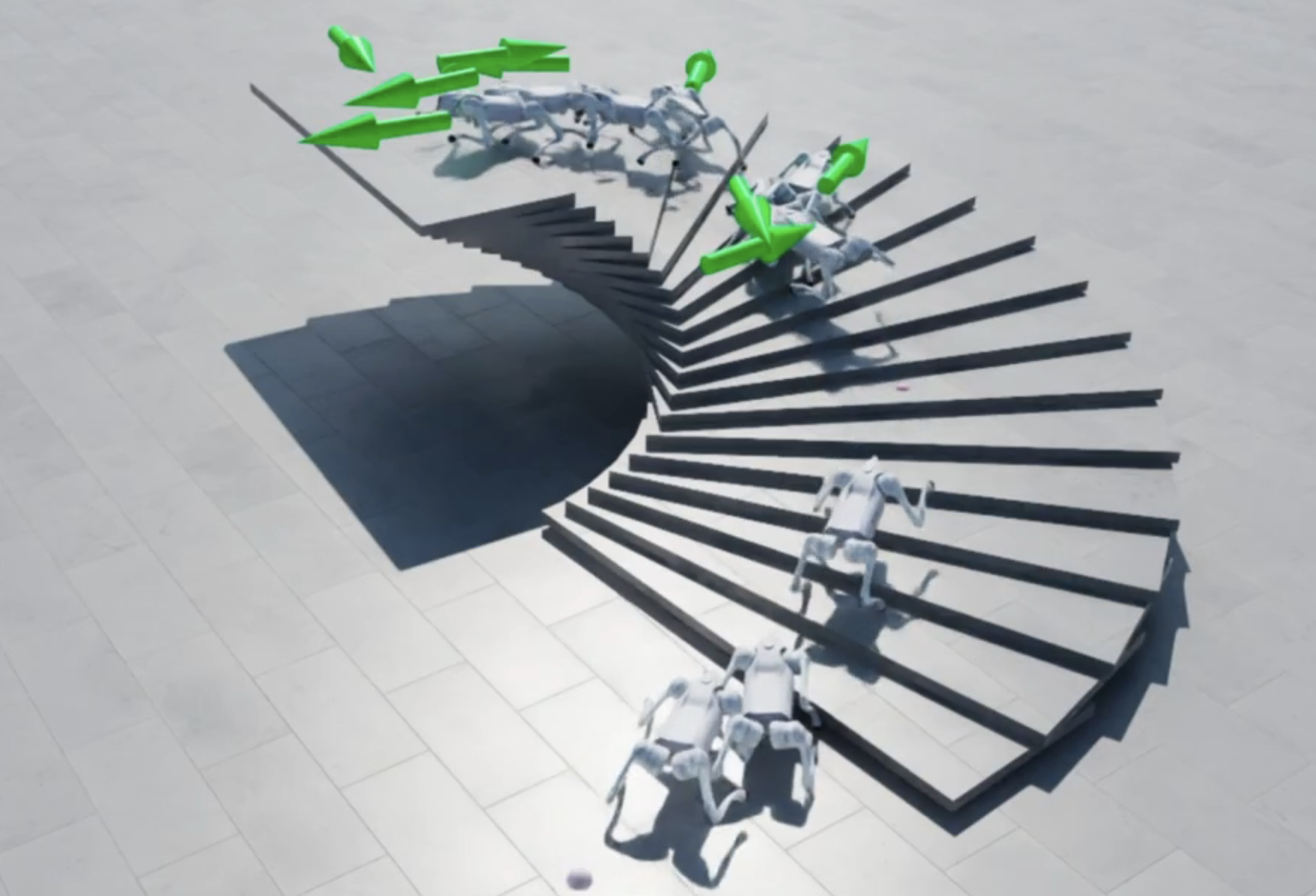}
        \caption{Robot Climbing Spiral Stairs}
        \label{fig:sub2}
    \end{subfigure}
    \caption{Robot Dogs Climbing Various Staircases}
    \label{fig_youtube}
\end{figure*}

\subsection{Regularization and penalties}

In addition to task rewards, there are regularization rewards that limit the quadruped from extreme behavior, such as moving too fast, creating jerky movement, or colliding with walls. Therefore, we presented noteworthy regularization rewards, while the regularization weights are left for the appendix.
\begin{equation}\label{eq_3}
r_{power} = -\sum_{i=1}^{12}\tau _{i}\dot{\theta _{i}}
\end{equation}
\begin{equation}\label{eq_4}
r_{stationary\ power} = -\sum_{i=1}^{12}\tau _{i}^{2}
\end{equation}
\begin{equation}\label{eq_5}
r_{action\ rate} = -\left\| a_{t} - a_{t-1}  \right\|_{2}^{2}
\end{equation}
\begin{equation}\label{eq_6}
r_{joint\ limit} = -\sum_{i=1}^{12}(ReLU(\theta _{i} - \theta _{max})+ ReLU(\theta _{min} - \theta _{i}))
\end{equation}
\begin{equation}\label{eq_7}
r_{joint\ velocity\ limit} = -\sum_{i=1}^{12}ReLU(\dot{\theta} - \dot{\theta} _{max})
\end{equation}
\begin{equation}\label{eq_8}
r_{joint\ velocity} = -\left\| \dot{\theta }  \right\|_{2}^{2}
\end{equation}
\begin{equation}\label{eq_9}
r_{joint\ acceleration} = -\left\| \ddot{\theta }  \right\|_{2}^{2}
\end{equation}

The equations ~\ref{eq_3}-~\ref{eq_9} above are joint-related regularization. $r_{power}$ and $r_{stationary\ power}$ penalize motor power output and excessive torque.  $r_{action\ rate}$ penalizes fast action change. $r_{joint\ limit}$ and $r_{joint\ velocity\ limit}$ penalize reaching the physical limit of the motors, whereareas $r_{joint\ velocity} $ and $r_{joint\ acceleration}$ penalize excessive joint velocity and acceleration.

We penalized collisions between the quadruped and objects. Specifically, a penalty of -8.0 is added when the quadruped’s body, head, hip, and thigh contact the ground or obstacles. An additional penalty of -0.2 is added when the calf is in contact, because mild contact with calf is acceptable.

To make the quadruped's gait more stable, we also penalized flying gait and encouraged larger steps through gait-related penalties and rewards. Denote $c_{i}(t) \in \left\{ 0,1\right\}$ as the foot $\mathit{i}$ contact state, which takes the value of 1 when in contact and 0 otherwise. The flying gait penalty and the step reward are $r_{flying} = \left\{\begin{matrix}-1 \ \forall \mathit{i}\in \left\{1,2,3,4 \right\} c_{i}(t)=0
 \\ 0 \ otherwise
\end{matrix}\right.$ and $r_{step} =  \sum_{i=1}^{4} 1- c_{i}(t)$. Finally, we penalized excessive action when the robot is in the goal region using the same reward function as $r_{action \ rate}$.

\subsection{Curriculum}

It was shown that using a terrain-based curriculum can assist the learning \cite{Lee_2020}; therefore, we employed our customized curriculum for the training of stair climbing. Each terrain type is divided into 10 levels, with each level progressively more difficult. In stage 1 training, pyramid stairs terrain increases its step height from 0 cm (flat terrain) to 12 cm. In stage 2 training, we increased the step height from 2cm to 12cm. In addition to step height, stair width also varies from 2.0 meters to 1.4 meters. One important feature of the curriculum design is that the length of the stairs increases as difficulty increases, and we observed that this design significantly improves learning speed. Specifically, for L-shaped terrain, the length of stairs after the turn ranges from 0 to 3 meters, which allows the policy to gradually learn to turn around the corner. In spiral terrain, the spiral length grows from 0.2 to 0.5 revolution as difficulty increases. The rule of leveling up follows Rudin et al \cite{rudin2022learningwalkminutesusing}. Given a quadruped at a specific level, it can level up to the more difficult terrain if it can reach the goal position with specified yaw. Otherwise, it gets demoted to a lower level if it fails to reach its goal. If it completes all levels, it is placed on a random level for continued training to prevent catastrophic forgetting of the policy.

\section{Experiments and Evaluations}
\label{sec:Experiments}

To test the performance of our resulting policy, we evaluated it on straight, L-shaped, and spiral stairs with parameters that differed from those in the training terrain. Given a single terrain type, we split the testing terrain into 6 levels, each with a different stair riser height ranging from 4cm to 14cm. The performance of the policy is measured by six metrics: goal-reached success rate, mean linear velocity, average climbing rate, position error, heading error, and mean power output.

\begin{itemize}[noitemsep]
    \item \textbf{Goal-Reached Success Rate} measures the percentage of quadrupeds that successfully reach the goal location under a single terrain type and difficulty level.
    \item \textbf{Mean Linear Velocity}, for each successful episode, measures the mean linear velocity over the whole testing episode. This informs us how agile the quadruped is as it completes the task.
    \item \textbf{Average Climbing Rate} is the velocity of a quadruped in the vertical direction over the whole testing episode. This informs us how quickly the quadruped ascends the stairs.
    \item \textbf{Position Error} is the distance error between the quadruped position and the commanded position. This informs us how accurately the quadruped reaches its goal.
    \item \textbf{Heading Error} is the angular error between the quadruped’s yaw and the commanded yaw. This also indicates how accurately the robot reaches its goal. 
    \item \textbf{Mean Power Output} is produced by the 12 motors averaged over the whole testing episode. This indicates the policy's efficiency.
\end{itemize}

\section{Results and Discussion}
\label{sec:Experiments}
\subsection{Model performance on different terrains}

In this testing, we rolled out 300 episodes and computed the mean metric values. At the beginning of the episode, the quadrupeds were placed 0.5 meters in front of the stairs with uniform [-0.3, 0.3] meters of random lateral offsets. The quadrupeds also had [-45, 45] degrees of random heading offset. Then, they were tasked to reach the target position and orientation randomly sampled from the terrain. During this process, metrics are collected and averaged for analysis.

\begin{figure}[!htb]
    \centering
    \includegraphics[width=0.4\textwidth]{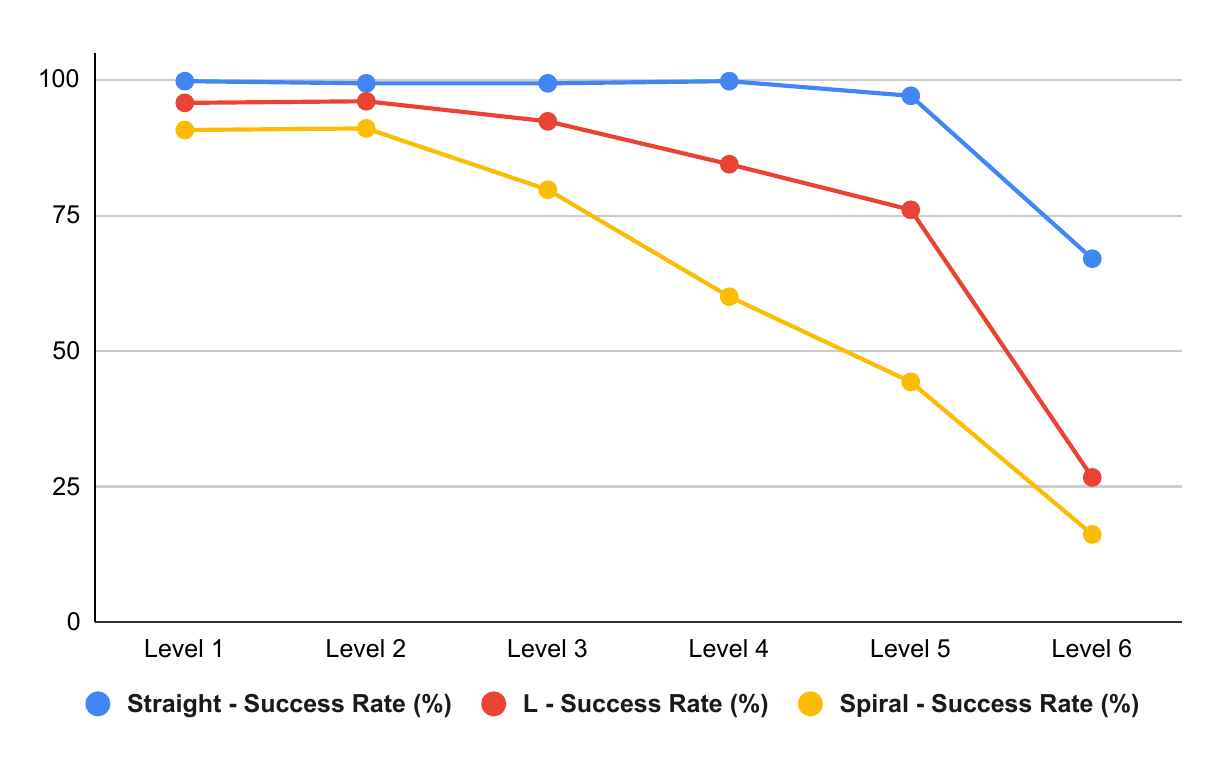}
    \caption{Levels vs. Success Rate (\%) for Straight, L-shaped, and Spiral Stairs}
    \label{fig_2}
\end{figure}

\begin{figure}[!htb]
    \centering
    \includegraphics[width=0.4\textwidth]{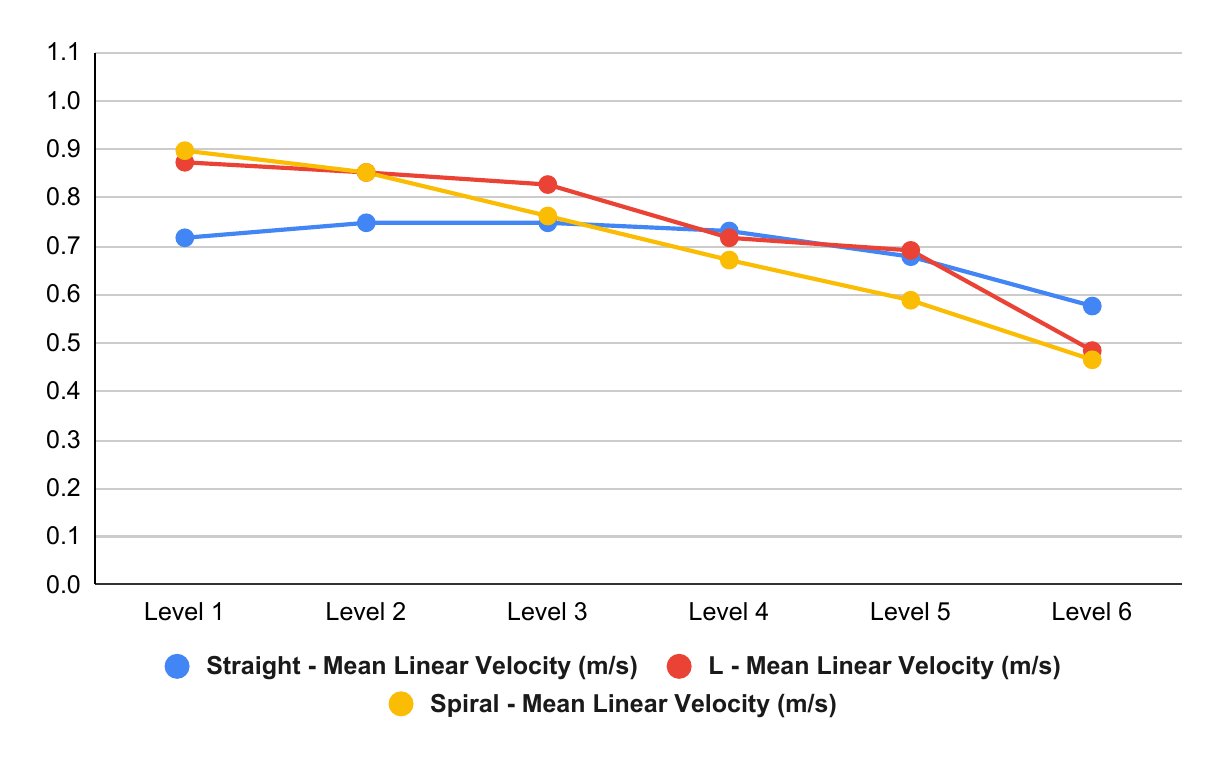}
    \caption{Levels vs. Linear Velocity (m/s) for Straight, L-shaped, and Spiral Stairs}
    \label{fig_3}
\end{figure}

\begin{figure}[!htb]
    \centering
    \includegraphics[width=0.4\textwidth]{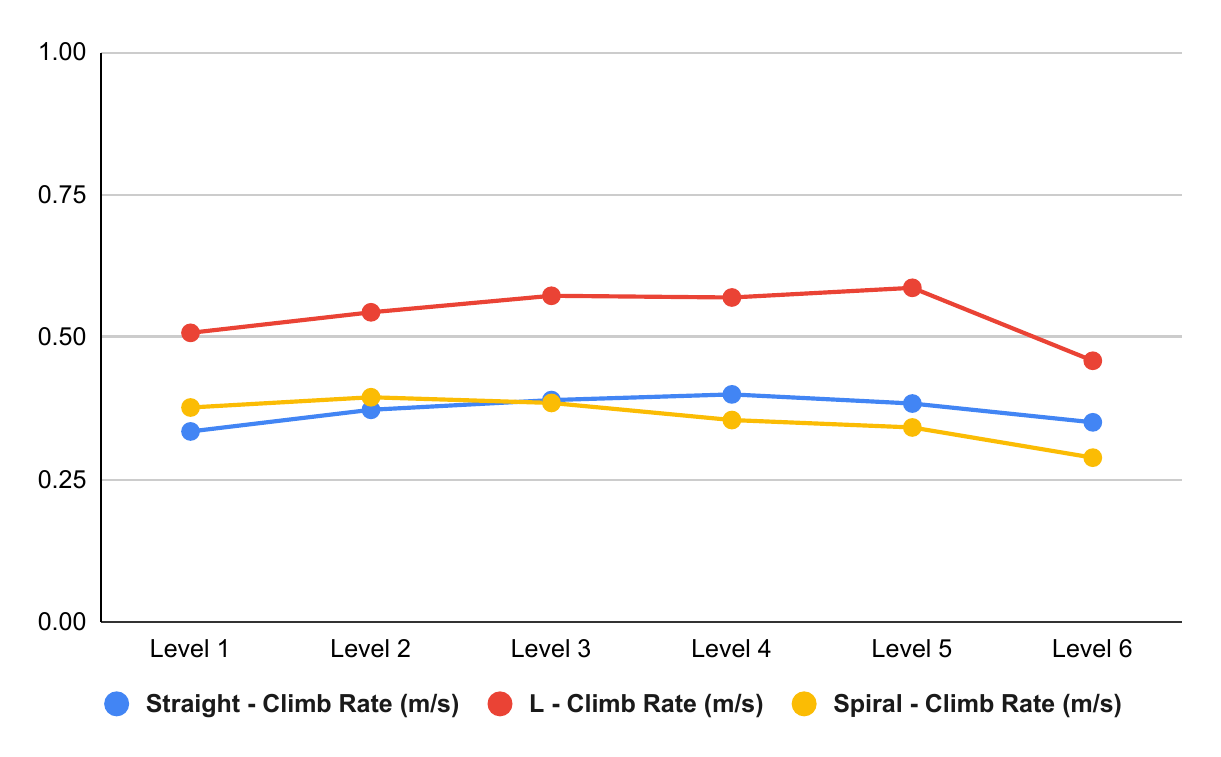}
    \caption{Levels vs. Climb Rate (m/s) for Straight, L-shaped, and Spiral Stairs}
    \label{fig_4}
\end{figure}

\begin{figure}[!htb]
    \centering
    \includegraphics[width=0.4\textwidth]{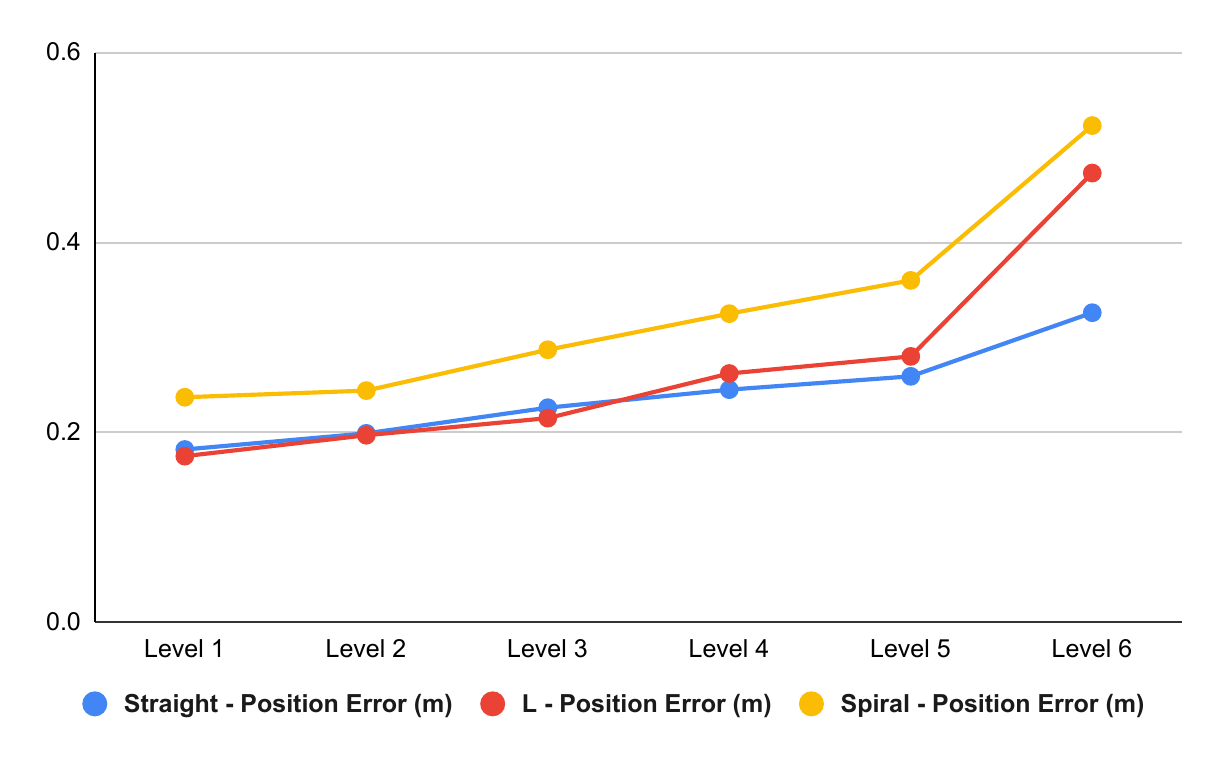}
    \caption{Levels vs. Position Error (m) for Straight, L-shaped, and Spiral Stairs}
    \label{fig_5}
\end{figure}

\begin{figure}[!htb]
    \centering
    \includegraphics[width=0.4\textwidth]{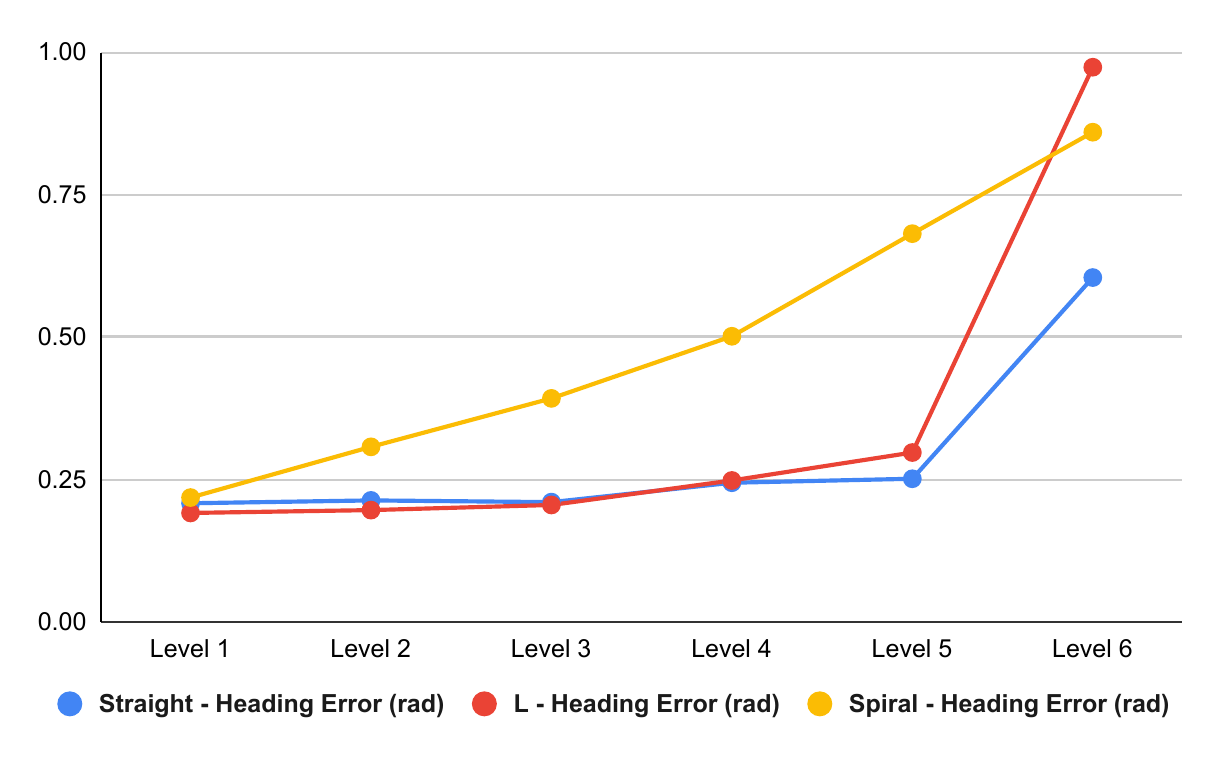}
    \caption{Levels vs. Heading Error (rad) for Straight, L-shaped, and Spiral Stairs}
    \label{fig_6}
\end{figure}

\begin{figure}[!htb]
    \centering
    \includegraphics[width=0.4\textwidth]{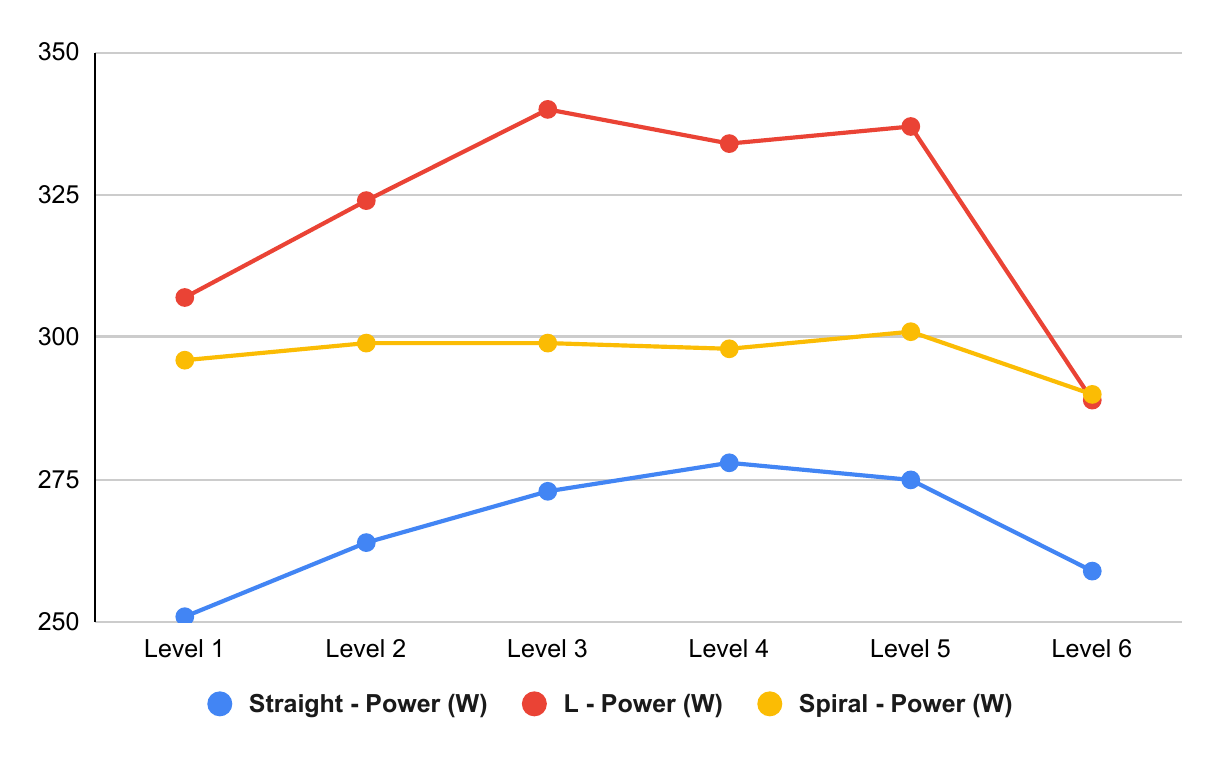}
    \caption{Levels vs. Power (W) for Straight, L-shaped, and Spiral Stairs}
    \label{fig_7}
\end{figure}

The testing results for three staircases, including straight, L-shaped, and Spiral, are recorded in a YouTube video \cite{youtube}. Figure \ref{fig_youtube} shows the YouTube screenshots of the results. Meanwhile, we comparatively demonstrated the performance of the quadrupeds in Figures ~\ref{fig_2} - \ref{fig_7}.

Figure \ref{fig_2} quantifies the reliability of climbing task completion as the experimental difficulty increases. The figure shows that higher levels are associated with a lower success rate, especially for spiral stairs. There is a greater decline in the success rate for robots climbing spiral stairs. This indicates that the spiral terrain is the most difficult terrain, followed by L-shaped terrain and straight terrain. Additionally, across all three stair types, there is a marked decline in success rate between the 5th and 6th levels. We observed that this is due to many quadrupeds hesitating to climb to the top floor at the 6th difficulty level. We hypothesize that attempting to climb higher stairs can lead to a high penalty due to a fall or collision, so the policy chooses to stay put and not attempt the climb.

Figures \ref{fig_3} and \ref{fig_4} show the robot's mean linear velocity and climb rate across different difficulty levels. Mean linear velocity generally decreases as level increases. Climb rate shows an overall increasing trend with level. It suggests that the robot's velocity and climb rate were less influenced by the complexity and difficulty of the stairs. 

As for the error handling, Figures \ref{fig_5} and \ref{fig_6} show that position and heading errors both increase with level. Robots climbing straight stairs have the smallest errors in both cases. Robots exhibit high position and heading errors in the L-shaped and Spiral staircases, particularly, there is a surge from the 5th to 6th level. This is expected, as many target poses are set on the stairs, it becomes increasingly difficult to align the quadruped's body to target poses on steep stairs.

As for power consumption, shown in Figure \ref{fig_7}, robots consume more power as the complexity and difficulty of the staircases increase. It should be noted that the decrease in power consumption at the 6th level could result from the policy adopting more conservative and slower motions to reduce the risk of failure, as aggressive behaviors will result in a penalty for failures.

\subsection{Comparing performance of two stages}
To assess the significance of two-stage training, we evaluate the policy trained on stage 1 on straight, L-shaped, and spiral terrains. By comparing its performance with terrain-specific models trained on stage 2, it quantifies the improvement in performance from second-stage training. All models are evaluated in level 3 difficulty terrains.

\begin{figure}
    \centering
    \includegraphics[width=0.9\linewidth]{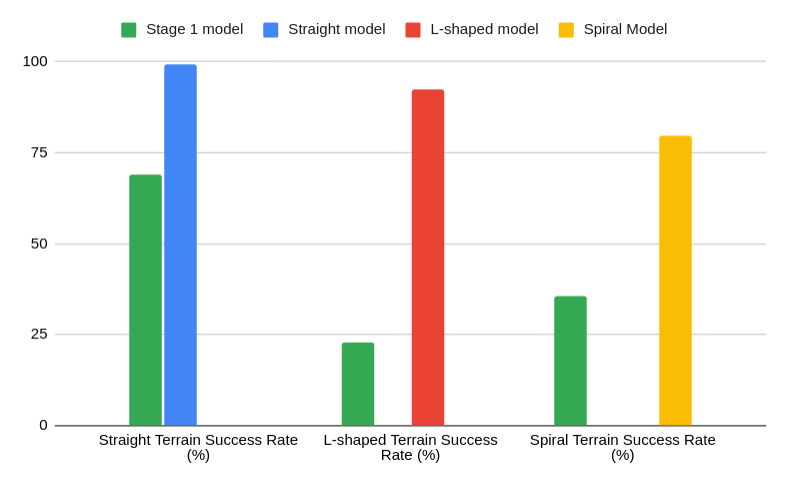}
    \caption{Terrains v.s. Success Rate. Terrains are level 3 difficulty.}
    \label{fig_9}
\end{figure}

Figure \ref{fig_9} shows the stage 1 model's success rate across three terrains, compared with other models specifically trained for those terrains. It is apparent that second-stage training can significantly improve model performance. In addition, it is noteworthy that the stage 1 model performs best on straight terrain, as it is the least difficult. The Stage 1 model performs worst on L-shaped terrain due to its inability to turn around the 90-degree corner. Rather, the quadruped 'short-cuts' the stairs and goes straight toward the goal position, resulting in a collision or fall. This indicates that the second stage of training enables the quadrupeds to recognize the shape of the stairs and navigate properly around the corner. We observed similar short-cutting behavior in spiral terrain, as the stage 1 model fails to follow the curvature of the spiral stairs.

\subsection{Discussion}
It should be noted that quadrupeds may have a lower success rate of reaching the top floor as difficulty increases, particularly when the staircases reach the 6th level; the success rate drops significantly. We replayed the climbing testing results and found that robot dogs usually hesitated to climb to the top floor at the 6th difficulty level. We speculated that the rewards of staying, as computed by the RL algorithm, are higher than those of taking risks to reach goals. Therefore, the robot chose conservative strategies. A potential strategy to mitigate this issue is to use exploration rewards, which encourage the robot to perform new behaviors. This would be the subject of our future research.

In our design, we use only local height map information rather than a global map. However, the results show that the robot can still navigate the stairs with this limited environmental information. This demonstrates the policy's robustness to partial observability, which is highly valued in a firefighting situation where a complete map of the building is unavailable.

Finally, it is important to note that the second-stage training is essential for acquiring navigation capability. The second-stage training, along with its CNN architecture, allows the model to recognize the stair patterns and the ascending direction. Without the stage 2 training, the model fails to recognize the direction of ascent from local terrain information and would choose a straight-line path toward the goal.

\subsection{Limitation}
In this project, we only tested ascending scenarios since it is the most straightforward firefight cases. After robot dogs climb to the top floors, they may be collected by firefighters. In the future, we will design and test more general algorithms for both ascending and descending cases. Additionally, we assumed the stairs were intact, but they may be damaged or obstructed during a fire. In the future, we will test our robot in a harsh environment.

\section{Conclusion}
\label{sec:Conclusion}
Quadrupeds are increasingly being studied and deployed for indoor primary search support because of their flexibility and efficiency. In this paper, we designed and tested an End-to-End two-stage RL approach that transfers stair-climbing skills from pyramided terrain to optimize both navigation and locomotion for a robot dog climbing various stairs, including straight, L-shaped, and spiral stairs. It shows a high success rate, high agility, and low position and heading errors with certain difficulty levels. Our RL algorithm design is with effective rewards, regularization, and penalties, and our curriculum strategies implement a cumulative learning process. Our centerline-based navigation formulation enables unified learning of navigation and locomotion without hierarchical planning. In the future, we will optimize our RL training to handle a wider range of climbing scenarios and test them on a real robot by deploying \textit{Isaac Lab}’s Sim-to-Real workflows. 
 
\bibliography{ISARC}

\end{document}